\documentclass[10pt,twocolumn,letterpaper]{article}

\usepackage{iccv}
\usepackage{times}
\usepackage{epsfig}
\usepackage{graphicx}
\usepackage{amsmath}
\usepackage{amssymb}

\usepackage{bm}
\usepackage{float}
\usepackage{threeparttable}
\usepackage{multirow}
\usepackage{booktabs}
\usepackage{array}
\usepackage{url}
\newcommand{\PreserveBackslash}[1]{\let\temp=\\#1\let\\=\temp}
\newcolumntype{C}[1]{>{\PreserveBackslash\centering}p{#1}}
\newcolumntype{R}[1]{>{\PreserveBackslash\raggedleft}p{#1}}
\newcolumntype{L}[1]{>{\PreserveBackslash\raggedright}p{#1}}
\renewcommand{\vec}[1]{\mathbf{#1}}
\newcommand{\matr}[1]{\bm{\mathit{#1}}}

\usepackage[pagebackref=true,breaklinks=true,letterpaper=true,colorlinks,bookmarks=false]{hyperref}

\iccvfinalcopy % *** Uncomment this line for the final submission

\def\iccvPaperID{5532} % *** Enter the ICCV Paper ID here

\ificcvfinal\fi
\begin{document}

%%%%%%%%% TITLE
\title{PR Product: A Substitute for Inner Product in Neural Networks}

\author{Zhennan Wang\footnotemark[2] , Wenbin Zou\footnotemark[2] , Chen Xu\footnotemark[1]\\
Shenzhen University\\
%Institution1 address\\
{\tt\small wangzhennan2017@email.szu.edu.cn, \{wzou, xuchen\_szu\}@szu.edu.cn}
% For a paper whose authors are all at the same institution,
% omit the following lines up until the closing ``}''.
% Additional authors and addresses can be added with ``\and'',
% just like the second author.
% To save space, use either the email address or home page, not both
%\and
%Second Author\\
%Institution2\\
%First line of institution2 address\\
%{\tt\small secondauthor@i2.org}
}
\maketitle
\thispagestyle{empty}

\footnotetext[2]{The authors are with College of Electronic and Information Engineering, the Shenzhen Key Laboratory of Advanced Machine Learning and Applications, the Guangdong Key Laboratory of Intelligent Information Processing, Shenzhen University.}
\footnotetext[1]{Corresponding author is with the College of Mathematics and Statistics, Shenzhen University.}

%%%%%%%%% ABSTRACT
\begin{abstract}
In this paper, we analyze the inner product of weight vector $\vec{w}$ and data vector $\vec{x}$ in neural networks from the perspective of vector orthogonal decomposition and prove that the direction gradient of $\vec{w}$ decreases with the angle between them close to 0 or $\pi$. We propose the Projection and Rejection Product (PR Product) to make the direction gradient of $\vec{w}$ independent of the angle and consistently larger than the one in standard inner product while keeping the forward propagation identical. As a reliable substitute for standard inner product, the PR Product can be applied into many existing deep learning modules, so we develop the PR Product version of fully connected layer, convolutional layer and LSTM layer. In static image classification, the experiments on CIFAR10 and CIFAR100 datasets demonstrate that the PR Product can robustly enhance the ability of various state-of-the-art classification networks. On the task of image captioning, even without any bells and whistles, our PR Product version of captioning model can compete or outperform the state-of-the-art models on MS COCO dataset. Code has been made available at:\url{https://github.com/wzn0828/PR_Product}.
\end{abstract}

%%%%%%%%% BODY TEXT
\section{Introduction}
% 参考 non-local neural networks;
% 整篇只讲1个故事；
% 以讲故事的语调来写，而不是干巴巴的罗列；
% 注重逻辑

% 1. 背景和related work
% 2. related work中存在的问题，最好list出来；（针对这个问题，之前的方法有什么不足；）我们提出了什么方法来解决这个问题，或我们的观察。
% 3. 这个问题为什么重要，可以举例说明；
% 4. 本文所提方法的介绍，可详可略，注重逻辑
% 5. 本文所提方法的优势，最好list出来
% 6. 应用本方法的实验以及实验结果（我们的实验中是怎么用本文提的方法的，定性的优势体现在哪里，最终结果的定量表现如何；）
% 7. a bullet list of contributions

%1.2.4.5.6.7必须有；2最重要。

\begin{figure}[t]
    \centering
    \includegraphics[width=\columnwidth]{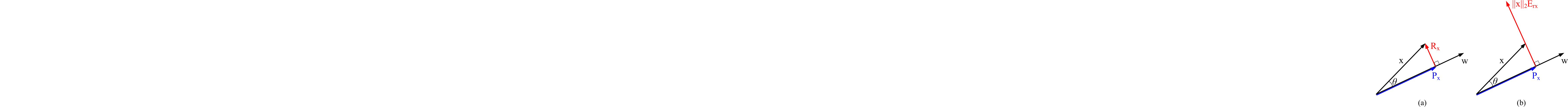}
    \caption{The orthogonal decomposition of the gradient w.r.t. weight vector $\vec{w}$ in two-dimensional space. (a) The case of the standard inner product. (b) The case of our proposed PR Product. For the length gradient, both are the vector projection $\Vec{P_x}$ of $\vec{x}$ onto $\vec{w}$. However, the direction gradient is changed from the vector rejection $\Vec{R_x}$ in (a) to $\|\vec{x}\|_2\vec{E_{rx}}$ in (b), where $\vec{E_{rx}}$ represents the unit vector along $\Vec{R_x}$. }
    \label{fig:pull_figure}
\end{figure}

% 1. 背景和related work: 神经网络取得了非凡的成就；但是其依赖基于梯度的学习；考虑一个最简单的操作wx， 仅仅把它看做加权求和，也就是代数计算，是简单的，也是很多研究者的共识。然而，在欧式空间中，从几何角度，他可以写成...
Models based on neural networks, especially deep convolutional neural networks (CNN) and recurrent neural networks (RNN), have achieved state-of-the-art results in various computer vision tasks~\cite{he2016deep,he2017mask,anderson2018bottom}. Most of the optimization algorithms for these models rely on gradient-based learning, so it is necessary to analyze the gradient of inner product between weight vector $ \vec{w} \in R^d $ and data vector $ \vec{x} \in R^d $, a basic operation in neural networks. Denoted by $P(\vec{w},\vec{x})= \vec{w}^T\vec{x}$ the inner product, the gradient of $P(\vec{w},\vec{x})$ w.r.t. $\vec{w}$ is exactly the data vector $\vec{x}$ which can be orthogonally decomposed into the vector projection $\Vec{P_x}$ on $\vec{w}$ and the vector rejection $\Vec{R_x}$ from $\vec{w}$, as shown in Figure \ref{fig:pull_figure} (a). The vector projection $\Vec{P_x}$ is parallel to the weight vector $\vec{w}$ and will update the length of $\vec{w}$ in next training iteration, called the length gradient. While the vector rejection $\Vec{R_x}$ is orthogonal to $\vec{w}$, it will change the direction of $\vec{w}$, called the direction gradient.
%Considering the inner product of weight vector $ w \in R^d $ and data vector $ x \in R^d $, a basic operation in neural networks, as a weighted summation operation is a general consensus, %which is exactly the algebraic definition of the inner product of two vectors. In this definition, the gradient of the inner product with respect to $\vec{w}$ is exactly the data vector $\vec{x}$.

% 2. related work中存在的问题，最好list出来；（针对这个问题，之前的方法有什么不足；）我们提出了什么方法来解决这个问题，或我们的观察。
Driven by the orthogonal decomposition of the gradient w.r.t. $\vec{w}$, a question arises: Which is the key factor for optimization, length gradient or direction gradient?
%To answer this question, we firstly optimize a 5-layer fully connected neural network with standard inner product on Fashion-MNIST~\cite{xiao2017fashion} and get an accuracy of 88.42\%. Then we replace the standard inner product with an variant, of which the length gradient w.r.t. $\vec{w}$ is zero, and get an accuracy of 88.32\%. And we also experiment with another variant, of which the direction gradient w.r.t. $\vec{w}$ is zero, and just get an accuracy of 38.59\%.
To answer this question, we optimize three 5-layer fully connected neural networks on Fashion-MNIST~\cite{xiao2017fashion} with different variants of inner product: standard inner product, a variant without length gradient and a variant without direction gradient. The top-1 accuracy is 88.42\%, 88.32\% and 38.59\%, respectively.
From these comparative experiments we can observe that the direction gradient is the key factor for optimization and is far more critical than the length gradient, which might be unsurprising. However, the direction gradient would be very small when $\vec{w}$ and $\vec{x}$ are nearly parallel, which would hamper the update of the direction of weight vector $\vec{w}$.

On the other hand, in Euclidean space, the geometric definition of inner product is the product of the Euclidean lengths of the two vectors and the cosine of the angle between them. That is $P(\vec{w},\vec{x})= \vec{w}^T\vec{x} = \|\vec{w}\|_2 \|\vec{x}\|_2 \cos \theta$, where we denote by $\|*\|_2$ the Euclidean length of vector $*$ and by $\theta$ the angle between $\vec{w}$ and $\vec{x}$ with the range of $[0, 2\pi)$.
%From this formulation, we can see that the $\theta$ has a significant impact on the dynamics of neural networks.
From this formulation, we can see that the $\theta$ is strongly connected with the direction of weight vector $\vec{w}$.
The gradient of $P$ w.r.t. $\theta$ is $\partial P / \partial \theta = -\|\vec{w}\|_2 \|\vec{x}\|_2 \sin \theta$, %which has several limitations during backpropagation. First, this gradient
which becomes small with $\theta$ close to 0 or $\pi$ and thus hinders the optimization. Several recent investigations of backpropagation ~\cite{bengio2013estimating,Yazdani2018LinearBI} focus on modifying the gradient of activation function. However, few researches propose variants of backpropagation for the inner product function.

%Second, as $\theta$ is the function of the unit vectors of $\vec{w}$ and $\vec{x}$, the smaller gradient w.r.t. $\theta$ will hamper the update of the direction of weight vector $\vec{w}$. Finally, it also discounts the direction gradient w.r.t. $\vec{x}$ and weakens the gradient flow to the downstream. As a result, the optimization becomes more and more difficult as the training of neural networks progresses. Several recent investigations of backpropagation ~\cite{bengio2013estimating,Yazdani2018LinearBI} focus on modifying the gradient of activation function. However, few researches propose variants of backpropagation of the inner product function.

% 4. 本文所提方法的介绍，可详可略，注重逻辑
In this paper, we propose the Projection and Rejection Product(abbreviated as PR Product) which changes the backpropagation of standard inner product to eliminate the dependence of the direction gradient of $\vec{w}$ and the gradient w.r.t. $\theta$ on the value of $\theta$. %, a substitute for inner product, which changes the backpropagation of inner product while keeping the same forward propagation.From the perspective of vector orthogonal decomposition, the vector $\vec{x}$ can be decomposed into vector projection $\Vec{P_x}$ on $\vec{w}$ and vector rejection $\Vec{R_x}$ from $\vec{w}$, as shown in Figure \ref{fig:pull_figure}.
We firstly prove that the standard inner product of $\vec{w}$ and $\vec{x}$ only contains the information of vector projection $\Vec{P_x}$, which is the main cause of the above dependence. While our proposed PR Product involves the information of both the vector projection $\Vec{P_x}$ and the vector rejection $\Vec{R_x}$ through rewriting the standard inner product into a different form and suitable components of that form are held fixed during backward pass. We further analyze the gradients of PR Product w.r.t. $\theta$ and $\vec{w}$. For $\theta$, the absolute value of gradient changes from  $\|\vec{w}\|_2 \|\vec{x}\|_2 \left|\sin \theta \right|$ to $\|\vec{w}\|_2 \|\vec{x}\|_2 $. For $\vec{w}$, the length of direction gradient changes from $\|\vec{x}\|_2\left|\sin\theta\right|$ to $\|\vec{x}\|_2$, as shown in Figure \ref{fig:pull_figure}.

%proving that the length of direction gradient of $\vec{w}$ is changed from the $\|\Vec{R_x}\|_2$ in standard inner product to $\|\vec{x}\|_2$ in PR Product, as shown in Figure \ref{fig:pull_figure}.

% 5. 本文所提方法的优势，最好list出来
There are several advantages of using PR Product:(a)The PR Product gets a different backward pass while the forward pass remains exactly the same as the standard inner product; (b) Compared with the behavior of standard inner product, %the length of direction gradient of $PR$ w.r.t. $\vec{w}$ is always larger and independent of $\theta$;
PR increases the proportion of the direction gradient which is the key factor for optimization; (c) As the PR Product maintains the linear property, it can be a reliable substitute for inner product operation in the fully connected layer, convolutional layer and recurrent layer. By reliable, we mean it does not introduce any additional parameters and matches with the original configurations such as activation function, batch normalization and dropout operation.

% 6. 应用本方法的实验以及实验结果（我们的实验中是怎么用本文提的方法的，定性的优势体现在哪里，最终结果的定量表现如何；）
We showcase the effectiveness of PR Product on image classification and image captioning tasks. For both tasks, we replace all the fully connected layers, convolutional layers and recurrent layers of the backbone models with their PR Product version. Experiments on image classification demonstrate that the PR Product can typically improve the accuracy of the state-of-the-art classification models. Moreover, our analysis on image captioning confirms that the PR Product definitely change the dynamics of neural networks. Without any tricks of improving the performance, like scene graph and ensemble strategy, our PR Product version of captioning model achieves results on par with the state-of-the-art models.

% 7. a bullet list of contributions
In summary, the main contributions of this paper are:
\begin{itemize}
\item We propose the PR Product, a reliable substitute for the standard inner product of weight vector $\vec{w}$ and data vector $\vec{x}$ in neural networks, which changes the
backpropagation while keeping the forward propagation identical;
\item We develop the PR-FC, PR-CNN and PR-LSTM, which applies the PR Product into the fully connected layer, convolutional layer and LSTM layer respectively;
\item Our experiments on image classification and image captioning suggest that the PR Product is generally effective and can become a basic operation of neural networks.
\end{itemize}

 \section{Related Work}
\noindent \textbf{Variants of Backpropagation.}
Several recent investigations have considered variants of standard backpropagation. In particular,~\cite{lillicrap2016random} presents a surprisingly simple backpropagation mechanism that assigns blame by multiplying errors signals with random weights, instead of the synaptic weights on each neuron, and further downstream.  ~\cite{baldi2016theory} exhaustively considers many Hebbian learning algorithms. The straight-through estimator proposed in ~\cite{bengio2013estimating} heuristically copies the gradient with respect to the stochastic output directly as an estimator of the gradient with respect to the sigmoid argument. ~\cite{Yazdani2018LinearBI} proposes Linear Backprop that backpropagates error terms only linearly. Different from these methods, our proposed PR Product changes the gradient of inner product function during backpropagation while maintaining the identical forward propagation.

\noindent \textbf{Image Classification.}
Deep convolutional neural networks~\cite{krizhevsky2014one,simonyan2014very,he2016deep,he2016identity,zagoruyko2016wide,xie2017aggregated,huang2017densely} have become the dominant machine learning approaches for image classification. To train very deep networks, shortcut connections have become an essential part of modern networks. For example, Highway Networks~\cite{srivastava2015highway,srivastava2015training} present shortcut connections with gating functions, while variants of ResNet~\cite{he2016deep,he2016identity,zagoruyko2016wide,xie2017aggregated} use identity shortcut connections. DenseNet~\cite{huang2017densely}, a more recent network with several parallel shortcut connections, connects each layer to every other layer in a feed-forward fashion.

\noindent \textbf{Image Captioning.}
In the early stage of vision to language field, template-based methods \cite{farhadi2010every,kulkarni2013babytalk} generate the caption templates whose slots are filled in by the outputs of object detection, attribute classification and scene recognition, which results in captions that sound unnatural. Recently, inspired by the advances in the NLP field, models based encoder-decoder architecture \cite{kiros2014multimodal,karpathy2015deep,wu2016value,jiang2018recurrent} have achieved striking advances. These approaches typically use a pretrained CNN model as the image encoder, combined with an RNN decoder trained to predict the probability distribution over a set of possible words.
%\cite{karpathy2015deep} adopts the result of object detection from R-CNN \cite{girshick2014rich} and a bidirectional RNN to learn a joint embedding space for caption ranking and generation. \cite{wu2016value} utilizes high-level concepts or attributes and inject them into a neural-based approach as semantic information to enhance image captioning system. \cite{jiang2018recurrent} proposes an intermediate fusion method which combines the representations of different image encoders to improve the performance.
To better incorporate the image information into the language processing, visual attention for image captioning was first introduced by \cite{xu2015show} which allows the decoder to automatically focus on the image subregions that are important for the current time step. Because of remarkable improvement of performance, many extensions of visual attention mechanism \cite{you2016image,chen2017sca,yang2016review,gu2017stack,lu2017knowing,anderson2018bottom} have been proposed to push the limits of this framework for caption generation tasks.
%\cite{you2016image} proposes a semantic attention model which selectively attends to semantic concept regions by fusing the global image representation and the semantic attributes representation from an attribute detector. \cite{chen2017sca} proposes a channel-wise attention model which takes advantage of the attention mechanism in feature maps of multiple layers and multiple channels. \cite{yang2016review} extends current attended encoder-decoder framework using a review network, which utilizes an attended LSTM to learn more compact and effective representations than those generated directly by the encoder. \cite{gu2017stack} proposes a stacked multi-stage caption prediction framework composed of multiple attended decoders each of which operates on the output of the previous stage, producing increasingly refined image descriptions. Rather than using the uniformly-divided grids of the outputs from CNN encoder as the attention units, \cite{anderson2018bottom} utilizes the object proposals of object detection as the basic attention units and applies attention mechanism on these proposals.
Except for those extensions to visual attention mechanism, several attempts \cite{rennie2017self,liu2017improved} have been made to adapt reinforcement learning to address the discrepancy between the training and the testing objectives for image captioning.
%Recently, the self-critical learning approach proposed by \cite{rennie2017self} utilizes the policy gradient algorithm and makes the reward of the sequence from the inference algorithm as a baseline to reduce the training variance. Such a training strategy can be leveraged to improve the performance of most of existing caption models under the encoder-decoder framework.
More recently, some methods ~\cite{yao2018exploring,johnson2018image,liu2018context,yang2018autoencoding} exploit scene graph to incorporate visual relationship knowledge into captioning models for better descriptive abilities.

\section{The Projection and Rejection Product}
In this section, we begin by shortly revisiting the standard inner product of weight vector $\vec{w}$ and data vector $\vec{x}$. % from the perspective of vector orthogonal decomposition.
Then we formally propose the Projection and Rejection Product (PR Product) which involves the information of both vector projection of $\vec{x}$ onto $\vec{w}$ and vector rejection of $\vec{x}$ from $\vec{w}$. Moreover, we analyze the gradient of PR Product. %w.r.t. weight vector $\vec{w}$.
Finally, %we show the implementation of PR Product and develop the PR-FC, PR-CNN, and PR-LSTM.
we develop the PR Product version of fully connected layer, convolutional layer and LSTM layer.
In the following, for the simplicity of derivation, we only consider a single data vector $\vec{x} \in R^d$ and a single weight vector $\vec{w} \in R^d$ except for the last subsection.

\subsection{Revisit the Inner Product in Neural Networks}
In Euclidean space, the inner product $P$ of the two Euclidean vectors $\vec{w}$ and $\vec{x}$ is defined by:
\begin{equation} \label{equ:innerproduct}
   P(\vec{w},\vec{x})= \vec{w}^T\vec{x} = \|\vec{w}\|_2 \|\vec{x}\|_2 \cos \theta
\end{equation}
where $\|*\|_2$ is the Euclidean length of vector $*$, and $\theta$ is the angle between $\vec{w}$ and $\vec{x}$ with the range of $[0, 2\pi)$. From this formulation, we can observe that the angle $\theta$ explicitly affects the state of neural networks.

\noindent\textbf{The gradient of \bm{$P$} w.r.t. \bm{$\theta$}.}
%The gradient of $P$ w.r.t. $\theta$.
Neither the weight vector $\vec{w}$ nor the data vector $\vec{x}$ is the function of $\theta$, so it is easy to get:
\begin{equation}\label{equ:conven_theta_grad}
    \frac{\partial P}{\partial \theta} = -\|\vec{w}\|_2 \|\vec{x}\|_2 \sin \theta
\end{equation}
%We argue that this is one of the reasons that the optimization becomes more and more difficult as the training progresses.

\noindent\textbf{The gradient of \bm{$P$} w.r.t. \bm{$\vec{w}$}.}
From Equation (\ref{equ:innerproduct}) and Figure \ref{fig:pull_figure} (a), it is easy to obtain the gradient function of $P$ w.r.t. $\vec{w}$:
\begin{equation}\label{equ:grad_P_w}
   \frac{\partial P}{\partial \vec{w}} = \vec{x} = \Vec{P_x} + \Vec{R_x}
\end{equation}
%The vector projection $\Vec{P_x}$ is parallel to the weight vector $\vec{w}$ and will update the length of $\vec{w}$ in next training iteration, called the length gradient of $\vec{w}$. While the vector rejection $\Vec{R_x}$ is orthogonal to $\vec{w}$, it will change the direction of $\vec{w}$, called the direction gradient. As the $\theta$ gets closer to 0 or $\pi$, the direction gradient $\Vec{R_x}$ becomes smaller and smaller, so it is increasingly difficult to update the direction of $\vec{w}$.

%As shown in Figure \ref{fig:pull_figure} (a), it is easy to get:
%\begin{equation}\label{equ:conven_w_grad}
%    \frac{\partial P}{\partial w} = x = \Vec{P_x} + \Vec{R_x}
%\end{equation}
%From the perspective of vector orthogonal decomposition, the vector $\vec{x}$ can be decomposed into vector projection on $\vec{w}$ and vector rejection from $\vec{w}$, as shown in Figure \ref{fig:pull_figure}. The former is the orthogonal projection of $\vec{x}$ onto $\vec{w}$, and the latter is the orthogonal projection of $x$ onto the hyperplane orthogonal to $\vec{w}$. We denote the vector projection of $x$ onto $\vec{w}$ by $\Vec{P_x}$ and the vector rejection of $x$ from $\vec{w}$ by $\Vec{R_x}$.
Here, $\Vec{R_x}$ is the direction gradient of $\vec{w}$. From Figure \ref{fig:pull_figure} (a) and Equation (\ref{equ:conven_theta_grad}), we can see that either the value of the gradient of $P$ w.r.t. $\theta$ or the length of $\Vec{R_x}$ is close to 0 with $\theta$ close to 0 or $\pi$, which would hamper the optimization of neural networks.

From Figure \ref{fig:pull_figure} (a),%Obviously, the length of $\Vec{P_x}$  is:
 we can easily get the length of $\Vec{P_x}$:
\begin{equation}\label{Px_length}
    \|\Vec{P_x}\|_2 = \|\vec{x}\|_2\left|\cos \theta \right|
\end{equation}
And the length of  $\Vec{R_x}$ is:
\begin{equation}\label{equ:Rx_length}
    \|\Vec{R_x}\|_2 = \|\vec{x}\|_2\left|\sin \theta \right|
\end{equation}
%From Figure \ref{fig:pull_figure} (a),
%According to Equation (\ref{Px_length}),
So equation (\ref{equ:innerproduct}) can be reformulated as:
\begin{equation}\label{equ:innerproduct_geo}
\begin{split}
   P(\vec{w},\vec{x}) &= \begin{cases}
              - \|\vec{w}\|_2\|\Vec{P_x}\|_2, & \mbox{if } \pi/2 \leq \theta < 3\pi/2;\\
               \|\vec{w}\|_2\|\Vec{P_x}\|_2, & \mbox{otherwise}.
             \end{cases} \\
     &= sign(\cos\theta)\|\vec{w}\|_2\|\Vec{P_x}\|_2
\end{split}
\end{equation}
where sign(*) denotes the sign of *. We can observe that this formulation only contains the information of vector projection of $\vec{x}$ on $\vec{w}$, $\Vec{P_x}$. As shown in Figure \ref{fig:pull_figure}, the vector projection $\Vec{P_x}$ changes very little when $\theta$ is near 0 or $\pi$, which may be a block to the optimization of neural networks. Although the length of the rejection vector $\Vec{R_x}$ is small when $\theta$ is close to 0 or $\pi$, it varies greatly  and thus is able to support the optimization of neural networks. That is our basic motivation for the proposed PR Product.

\subsection{The PR Product}
In order to take advantage of the vector rejection, the simplest way is to replace the $\|\Vec{P_x}\|_2$ in Equation (\ref{equ:innerproduct_geo}) with $\|\Vec{P_x}\|_2 + \|\Vec{R_x}\|_2$. But the trends of $\|\Vec{P_x}\|_2$ and $\|\Vec{R_x}\|_2$ with $\theta$ are inconsistent, so we employ $\|\vec{x}\|_2 - \|\Vec{R_x}\|_2$ to involve the information of vector rejection. In addition, we utilize two coefficients to maintain the linear property, which are held fixed during the backward pass.
% To take advantage of both the vector projection $\Vec{P_x}$ and the vector rejection $\Vec{R_x}$ while maintaining the same forward pass, we reformulate the inner product of $\vec{w}$ and $\vec{x}$ into the following form:
To be more detailed, we derive the PR Product as follows:
\begin{small}
\begin{equation}\label{equ:PR-Net}
\begin{split}
   &PR(\vec{w},\vec{x}) \\
   =&sign(\cos\theta)\|\vec{w}\|_2\left[\underline{\dfrac{\|\Vec{R_x}\|_2}{\|\vec{x}\|_2}}\|\Vec{P_x}\|_2 +\underline{\dfrac{\|\Vec{P_x}\|_2}{\|\vec{x}\|_2}}(\|\vec{x}\|_2 - \|\Vec{R_x}\|_2)\right]  \\
   =&\|\vec{w}\|_2\left[\underline{\left|\sin \theta\right|}\|\Vec{P_x}\|_2sign(\cos \theta) + \underline{\cos \theta}(\|\vec{x}\|_2-\|\Vec{R_x}\|_2)\right]\\
   =&\|\vec{w}\|_2 \|\vec{x}\|_2 \left[\underline{\left|\sin \theta\right|} \cos \theta +  \underline{\cos \theta}(1-\left | \sin \theta \right |)  \right]
\end{split}
\end{equation}
\end{small}
For clarity, we denote by $PR$ the proposed product function. Note that the \underline{*} denotes detaching * from neural networks. By detaching, we mean * is considered as a constant rather than a variable during backward propagation. Compared with the standard inner product formulation (Equation (\ref{equ:innerproduct_geo}) or (\ref{equ:innerproduct})), this formulation involves not only the information of vector projection $\Vec{P_x}$ but also the one of vector rejection $\Vec{R_x}$ without any additional parameters. We call this formulation the Projection and Rejection Product or PR Product for brevity.

%The PR Product keeps the same forward propagation as the standard inner product, which means it maintains the linear property.
Although the PR Product does not change the outcome during forward pass, compared with the standard inner product, it changes the gradients during backward pass. In the following, we theoretically derive the gradient of $PR$ w.r.t. $\theta$ and $\vec{w}$ during backpropagation.

\noindent\textbf{The gradient of \bm{$PR$} w.r.t. \bm{$\theta$}.}
%From Figure \ref{fig:pull_figure}, we can see that neither the weight vector $\vec{w}$ nor the data vector $\vec{x}$ is the function of $\theta$. So
We just need to calculate the gradients of trigonometric functions except for the detached ones in Equation (\ref{equ:PR-Net}). When $\theta$ is in the range of $[0, \pi)$, the gradient of $PR$ w.r.t. $\theta$ is:
\begin{equation}
\begin{split}
   \frac{\partial PR}{\partial \theta} &= \|\vec{w}\|_2 \|\vec{x}\|_2\left( -\sin^2 \theta - \cos^2 \theta \right) \\
     &= -\|\vec{w}\|_2 \|\vec{x}\|_2
\end{split}
\end{equation}
When $\theta$ is in the range of $[\pi, 2\pi)$, the gradient of $PR$ w.r.t. $\theta$ is:
\begin{equation}
\begin{split}
   \frac{\partial PR}{\partial \theta} &= \|\vec{w}\|_2 \|\vec{x}\|_2\left( \sin^2 \theta + \cos^2 \theta \right) \\
     &= \|\vec{w}\|_2 \|\vec{x}\|_2
\end{split}
\end{equation}
We use the following unified form to express the above two cases:
\begin{equation}\label{equ:PR_grad_theta}
    \frac{\partial PR}{\partial \theta} = \|\vec{w}\|_2 \|\vec{x}\|_2 sign\left(- \sin \theta\right)
\end{equation}

Compared with the standard inner product (Equation (\ref{equ:conven_theta_grad})), the PR Product changes the gradient w.r.t. $\theta$ from a smoothing function to a hard one. One advantage of this is the gradient w.r.t. $\theta$ does not decrease as $\theta$ gets close to 0 or $\pi$, providing continuous power for the optimization of neural networks.

\noindent\textbf{The gradient of \bm{$PR$} w.r.t. \bm{$\vec{w}$}.}
Above we discussed the gradient w.r.t. $\theta$, an implicit variable in neural networks. In this part, we explicitly take a look at the differences between the gradients of the standard inner product and our proposed PR Product w.r.t. $\vec{w}$ .

%We first analyze the gradient of standard inner product w.r.t. $\vec{w}$. From Equation (\ref{equ:innerproduct}) and Figure \ref{fig:pull_figure}, it is easy to obtain the gradient function of $P$ w.r.t.  $\vec{w}$:
%\begin{equation}\label{equ:grad_P_w}
%   \frac{\partial P}{\partial w} = x = \Vec{P_x} + \Vec{R_x}
%\end{equation}
%The vector projection $\Vec{P_x}$ is parallel to the weight vector $\vec{w}$ and will update the length of $\vec{w}$ in next training iteration, called the length gradient of $\vec{w}$. While the vector rejection $\Vec{R_x}$ is orthogonal to $\vec{w}$, it will change the direction of $\vec{w}$, called the direction gradient. As the $\theta$ gets closer to 0 or $\pi$, the direction gradient $\Vec{R_x}$ becomes smaller and smaller, so it is increasingly difficult to update the direction of $\vec{w}$.

For the PR Product, we derive the gradient of $PR$ w.r.t. $\vec{w}$ from Equation (\ref{equ:PR-Net}) and Equation (\ref{equ:PR_grad_theta}) as follows :
\begin{equation}\label{equ:grad_PR_w}
  \begin{split}
   &\frac{\partial PR}{\partial \vec{w}}\\
  =& \frac{\vec{w}}{\|\vec{w}\|_2}\|\vec{x}\|_2\cos \theta + \|\vec{w}\|_2\|\vec{x}\|_2sign(-\sin\theta)\frac{\partial \theta}{\partial \vec{w}}\\
  =& \vec{P_x} + \|\vec{w}\|_2\|\vec{x}\|_2sign(-\sin\theta)\frac{\mathrm{d} \theta}{\mathrm{d}\cos \theta} \frac{\partial \cos \theta}{\partial \vec{w}}\\
  =& \vec{P_x} + \frac{\|\vec{w}\|_2\|\vec{x}\|_2}{|\sin \theta|}\frac{\partial\left(\frac {\vec{w}^T\vec{x}}{\|\vec{w}\|_2\|\vec{x}\|_2}\right)}{\partial \vec{w}}\\
  =& \vec{P_x} + \frac{\|\vec{w}\|_2\|\vec{x}\|_2}{|\sin \theta|}\frac{(\matr{I}-\matr{M_w})\vec{x}}{\|\vec{w}\|_2\|\vec{x}\|_2}\\
  =& \vec{P_x} + \frac{\vec{R_x}}{|\sin\theta|}\\
  =& \vec{P_x} + \|\vec{x}\|_2\frac{\vec{R_x}}{\|\vec{R_x}\|_2}\\
  =& \vec{P_x} + \|\vec{x}\|_2\vec{E_{rx}}, \qquad with \quad \matr{M_w}=\frac{\vec{w}\vec{w}^T}{\| \vec{w} \|_2^2}
  \end{split}
\end{equation}
Where $\matr{M_w}$ is the projection matrix that projects onto the weight vector $\vec{w}$, which means $\matr{M_w}\vec{x}=\vec{P_x}$, and $\vec{E_{rx}}$ is the unit vector along the vector rejection $\vec{R_x}$. Similar to Equation (\ref{equ:grad_P_w}), the $\vec{P_x}$ is the length gradient part and the $\|\vec{x}\|_2\vec{E_{rx}}$ is the direction gradient part. For the length gradient, the cases in $P$ and $PR$ are identical. For the direction gradient part, however, the one in $PR$ is consistently larger than the one in $P$, except for the almost impossible cases when $\theta$ is equal to $\pi/2$ or $3\pi/2$. So $PR$ increases the proportion of the direction gradient. In addition, the length of direction gradient in $PR$ is independent of the value of $\theta$. Figure \ref{fig:pull_figure} shows the comparison of the gradients of the two formulations w.r.t. $\vec{w}$.

%\subsection{Implementation of PR Product}
%As mentioned above, $\theta$ is an implicit variable in neural networks, so we can't directly implement the PR Product according to Equation (\ref{equ:PR-Net}). By Equation (\ref{equ:Rx_length}) and the Pythagorean theorem, we can derive $|\sin\theta|$ as follows:
%\begin{equation}\label{equ:sin_abs}
%\begin{split}
%   |\sin \theta| &= \frac{\|\Vec{R_x}\|_2}{\|\vec{x}\|_2} \\
%     &= \frac{\sqrt{\|\vec{x}\|_2^2-\|\Vec{P_x}\|_2^2}}{\|\vec{x}\|_2}\\
%     &=\frac{\sqrt{\|\vec{w}\|_2^2\|\vec{x}\|_2^2-(\vec{w}^T\vec{x})^2}}{\|\vec{w}\|_2\|\vec{x}\|_2}
%\end{split}
%\end{equation}
%Substituting it into Equation (\ref{equ:PR-Net}), we can get the implementation of PR Product in practice:
%\begin{equation}\label{equ:PR-Net_imple}
%  \begin{split}
%     &PR(\vec{w},\vec{x})= \underline{\frac{\sqrt{\|\vec{w}\|_2^2\|\vec{x}\|_2^2-(\vec{w}^T\vec{x})^2}}{\|\vec{w}\|_2\|\vec{x}\|_2}}\vec{w}^T\vec{x} \quad + \\
%     &\underline{\frac{\vec{w}^T\vec{x}}{\|\vec{w}\|_2\|\vec{x}\|_2}}\left( \|\vec{w}\|_2\|\vec{x}\|_2 - \sqrt{\|\vec{w}\|_2^2\|\vec{x}\|_2^2-(\vec{w}^T\vec{x})^2} \right)
%  \end{split}
%\end{equation}
%Also, the \underline{*} denotes detaching * from neural networks.

\subsection{PR-X}
The PR Product is a reliable substitute for the standard inner product operation, so it can be applied into many existing deep learning modules, such as fully connected layer(FC), convolutional layer(CNN) and LSTM layer. We denote the module X with PR Product by PR-X. In this section, we show the implementation of PR-FC, PR-CNN and PR-LSTM.

\noindent\textbf{PR-FC.}
To get PR-FC, we just replace the inner product of the data vector $\vec{x}$ and each weight vector in the weight matrix with the PR Product. Suppose the weight matrix $\matr{W}$ contains a set of n column vectors, $\matr{W}=(\vec{w_1},\vec{w_2},...,\vec{w_n})$, so the output vector of PR-FC can be calculated as follows:
\begin{equation}\label{equ:PR_FC}
\begin{split}
    &PR\text{-}FC(\matr{W},\vec{x})\\
   =& \left( PR(\vec{w_1},\vec{x}) , PR(\vec{w_2},\vec{x}),...,PR(\vec{w_n},\vec{x}) \right) + \vec{b}
\end{split}
\end{equation}
where $\vec{b}$ represents an additive bias vector if any.

\noindent\textbf{PR-CNN.}
To apply the PR Product into CNN, we convert the weight tensor of the convolutional kernel and the data tensor in the sliding window into vectors in Euclidean space, and then use the PR Product to calculate the output. Suppose the size of the convolution kernel $\vec{w}$ is $(k_1, k_2, C_{in})$, so the output at position (i, j) is:
\begin{equation}\label{equ:PR_CNN}
\begin{split}
  &PR\text{-}CNN(\vec{w},\vec{x})_{ij}\\
  =&PR(flatten(\vec{w}), flatten(\vec{x_{[ij]}})) + b
\end{split}
\end{equation}
where $flatten(\vec{w})$ and $flatten(\vec{x_{[ij]}})$ $\in R^{k_1*k_2*C_{in}}$, $\vec{x_{[ij]}}$ represents the data tensor in the sliding window corresponding to output position (i,j), and $b$ represents an additive bias if any.

\noindent\textbf{PR-LSTM.}
To get the PR Product version of LSTM, just replace all the perceptrons in each gate function with the PR-FC.
For each element in input sequence, each layer computes the following function:
\begin{small}
\begin{equation}
\begin{split}
    &\vec{i_t} = \sigma\left(PR\text{-}FC(\matr{W_{ii}}, \vec{x_t}) + PR\text{-}FC(\matr{W_{hi}}, \vec{h_{(t-1)}}) + \vec{b_i}\right) \\
    &\vec{f_t} = \sigma\left(PR\text{-}FC(\matr{W_{if}}, \vec{x_t}) + PR\text{-}FC(\matr{W_{hf}}, \vec{h_{(t-1)}}) + \vec{b_f}\right) \\
    &\vec{g_t} = \tanh\left(PR\text{-}FC(\matr{W_{ig}}, \vec{x_t}) + PR\text{-}FC(\matr{W_{hg}}, \vec{h_{(t-1)}}) + \vec{b_g}\right) \\
    &\vec{o_t} = \sigma\left(PR\text{-}FC(\matr{W_{io}}, \vec{x_t}) + PR\text{-}FC(\matr{W_{ho}}, \vec{h_{(t-1)}}) + \vec{b_o}\right ) \\
    &\vec{c_t} = \vec{f_t}*\vec{c_{(t-1)}} + \vec{i_t}*\vec{g_t} \\
    &\vec{h_t} = \vec{o_t}*\tanh(\vec{c_t})
\end{split}
\end{equation}
\end{small}
%where $\vec{h_t}$ is the hidden state at time t, $\vec{c_t}$ is the cell state at time t, $\vec{x_t}$ is the input at time t, $\vec{h_{(t-1)}}$ is the hidden state of the layer at time t-1 or the initial hidden state at time 0, and $\vec{i_t}$, $\vec{f_t}$, $\vec{g_t}$, $\vec{o_t}$ are the input, forget, cell, and output gates, respectively. $\sigma$ is the sigmoid function, and * is the Hadamard product.
where $\sigma$ is the sigmoid function, and * is the Hadamard product.

In the following, we conduct experiments on image classification to validate the effectiveness of PR-CNN. And then we show the effectiveness of PR-FC and PR-LSTM on image captioning task.
%To validate the effectiveness of PR-CNN, we conduct experiments on varied applications including image captioning and image classification in next section.
\begin{table}[t]
%\begin{table}
  \centering
  \begin{threeparttable}
    %\begin{tabular}{ccccccc}
    \begin{tabular*}{\hsize}{@{}@{\extracolsep{\fill}}lcccccccccccccccc@{}}
    \toprule
    %\multirow{1}{*}{Modle}&
    %\multicolumn{8}{c}{Cross-Entropy Loss}\cr
    %\cmidrule(lr){1-9}
    \multicolumn{1}{c}{Model}                &CIFAR10         &CIFAR100\cr
    \midrule
    ResNet110               &6.23          &28.08\cr
    PR-ResNet110             &\textbf{5.97}          &\textbf{27.88}\cr
    \hline
    PreResNet110       &5.99          &27.08\cr
    PR-PreResNet110    &\textbf{5.64}          &\textbf{26.82}\cr
    \hline
    WRN-28-10              &4.34          &\textbf{19.50}\cr
    PR-WRN-28-10         &\textbf{4.03}             &19.57\cr
    \hline
    DenseNet-BC-100-12        &4.63             &22.88 \cr
    PR-DenseNet-BC-100-12     &\textbf{4.46}    &\textbf{22.64} \cr
    \bottomrule
    \end{tabular*}
  \end{threeparttable}
  \caption{Error rates on CIFAR10 and CIFAR100. The best results are highlighted in bold for the models with the same backbone architectures. All values are reported in percentage. The PR Product version can typically outperform the corresponding backbone models.}
  \label{tab:comparison on CIFAR}
\end{table}

\section{Experiments on Image Classification}
\subsection{Classification Models}
We employ various classic networks such as ResNet~\cite{he2016deep}, PreResNet~\cite{he2016identity}, WideResNet~\cite{zagoruyko2016wide} and DenseNet-BC~\cite{huang2017densely} as the backbone networks in our experiments. In particular, we consider ResNet with 110 layers denoted by ResNet110, PreResNet with 110 layers denoted by PreResNet110, and WideResNet with 28 layers and a widen factor of 10 denoted by WRN-28-10, as well as DenseNet-BC with 100 layers and a growth rate of 12 denoted by DenseNet-BC-100-12. For ResNet110 and PreResNet110, we use the classic basic block. To get the corresponding PR Product version models, all the fully connected layers and the convolutional layers in the backbone models are replaced with our PR-FC and PR-CNN respectively, and we denote them by PR-X, such as PR-ResNet110, PR-PreResNet110, PR-WRN-28-10 and PR-DenseNet-BC-100-12 respectively.
\begin{figure}[t]
    \centering
    \includegraphics[width=\columnwidth]{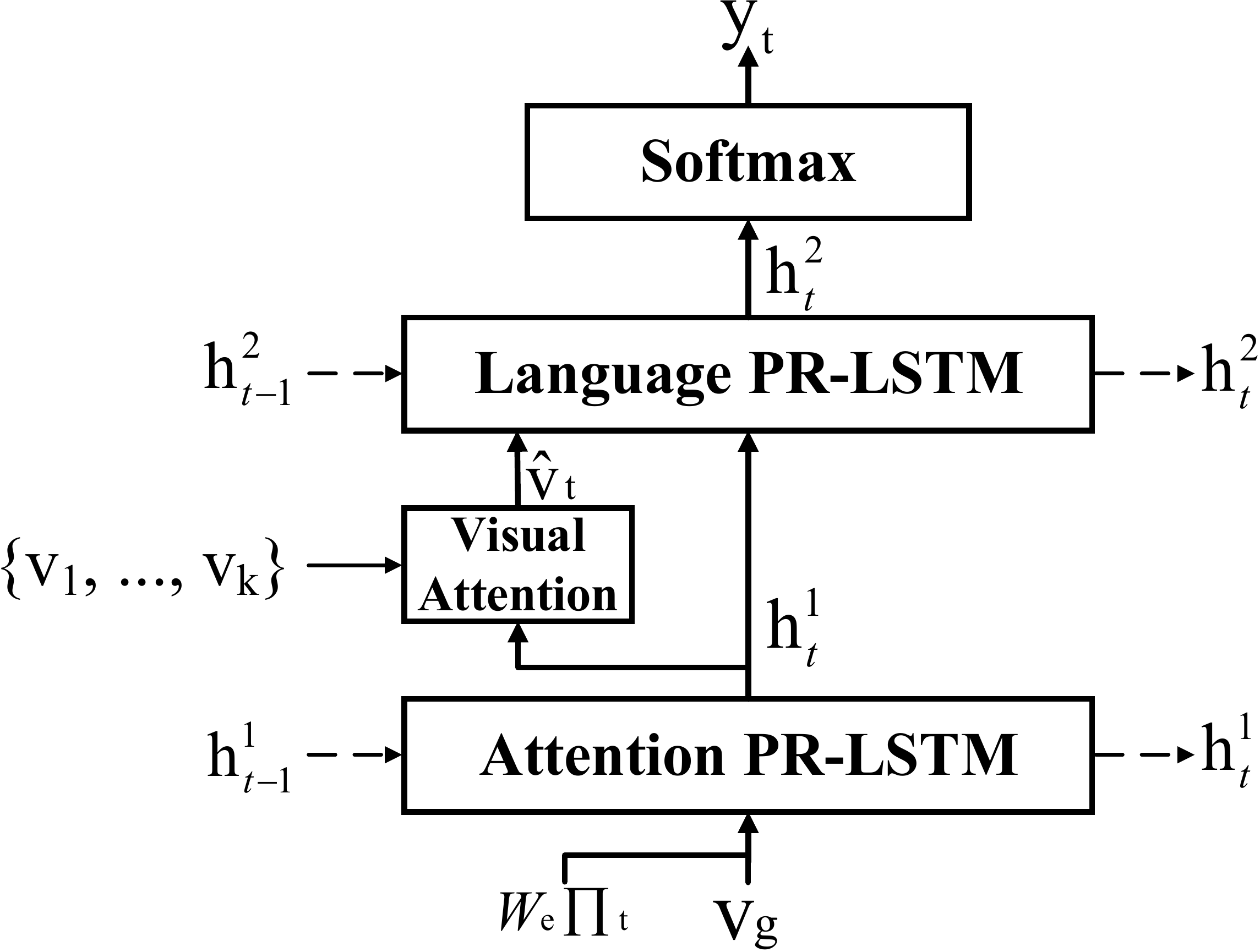}
    \caption{Decoder module used in our captioning model. The input to the Attention PR-LSTM consists of the global image representation $\vec{v_g}$ and the embedding of the previously generated word $\matr{W_e}\vec{\Pi_t}$. The input to the Language PR-LSTM consists of the attended image representation $\vec{\hat{v}_t}$ concatenated with the output of the Attention PR-LSTM. The dotted arrows represent the transfer of the hidden states of PR-LSTM layers.}
    \label{fig:caption_decoder}
\end{figure}

\subsection{Dataset and Settings}
We conduct our image classification experiments on the CIFAR dataset~\cite{krizhevsky2009learning}, which consists of 50k and 10k images of $32 \times 32$ pixels for the training and test sets respectively. The images are labeled with 10 and 100 categories, namely CIFAR10 and CIFAR100 datasets. We present experiments trained on the training set and evaluated on the test set. We follow the simple data augmentation in~\cite{lee2015deeply} for training: 4 pixels are padded on each side and a $32\times32$ crop is randomly sampled from the padded image or its horizontal flip. For testing, we only evaluate the single view of the original $32\times32$ image. Note that our focus is on the effectiveness of our proposed PR Product, not on pushing the state-of-the-art results, so we do not use any more data augmentation and training tricks to improve accuracy.

\subsection{Results and Analysis}
For fair comparison, not only are the PR-X models trained from scratch but also the corresponding backbone models, so our results may be slightly different from the ones presented in the original papers due to some hyper-parameters like random number seeds. The strategies and hyper-parameters used to train the respective backbone models, such as the optimization solver, learning rate schedule, parameter initialization method, random seed for initialization, batch size and weight decay, are adopted to train the corresponding PR-X models. The results are shown in Table \ref{tab:comparison on CIFAR} and some training curves are shown in the supplementary material, from which we can see that the PR-X can typically improve the corresponding backbone models on both CIFAR10 and CIFAR100. On average, it reduces the top-1 error by 0.27\% on CIFAR10 and 0.16\% on CIFAR100. It is worth emphasizing that the PR-X models don't introduce any additional parameters and keep the same hyper-parameters as the corresponding backbone models.
\begin{table}[t]
%\begin{table}
  \centering
  \begin{threeparttable}
    %\begin{tabular}{ccccccc}
    \begin{tabular*}{\columnwidth}{@{\extracolsep{\fill}}L{1.1cm}C{0.4cm}C{0.4cm}C{0.4cm}C{0.4cm}C{0.4cm}C{0.4cm}C{0.5cm}C{0.4cm}}
    \toprule
    %\multirow{1}{*}{Modle}&
    %\multicolumn{8}{c}{Cross-Entropy Loss}\cr
    %\cmidrule(lr){1-9}
    Product&B1&B2&B3&B4&M&RL&C&S\cr
    \midrule
    P&76.7&60.8&47.3&36.8&28.1&56.9&116.0&\textbf{21.1}\cr
    R&76.3&60.4&46.7&36.0&27.7&56.5&113.3&20.6\cr
    PR&\textbf{76.8}&\textbf{61.0}&\textbf{47.5}&\textbf{37.0}&\textbf{28.2}&\textbf{57.1}&\textbf{116.1}&\textbf{21.1} \cr
    \hline
    P$^*$&80.3&\textbf{64.9}&50.4&38.6&28.6&58.4&127.2&\textbf{22.4}\cr
    R$^*$&80.0&64.6&49.8&37.6&28.3&57.8&125.5&22.0\cr
    PR$^*$&\textbf{80.4}&\textbf{64.9}&\textbf{50.5}&\textbf{38.7}&\textbf{28.8}&\textbf{58.5}&\textbf{128.3}&\textbf{22.4}\cr
    \bottomrule
    \end{tabular*}
  \end{threeparttable}
  \caption{Performance comparison of different products on the test portion of Karpathy splits on MS COCO dataset, where Bn is short for BLEU-n, M is short for METEOR, RL is short for ROUGE-L, C is short for CIDEr, and S is short for SPICE. The top part is for cross-entropy training, and the bottom part is for CIDEr optimization (marked with $^*$). All values are reported in percentage, with the highest value of each entry highlighted in boldface.}
  \label{tab:model variants_comparison}
\end{table}

\section{Experiments on Image Captioning}
\subsection{Captioning Model}
We utilize the widely used encoder-decoder framework \cite{anderson2018bottom,lu2017knowing} as our backbone model for image captioning.

\noindent\textbf{Encoder.}
We use the Bottom-Up model proposed in \cite{anderson2018bottom} to generate the regional representations and the global representation of a given image $\matr{I}$. The Bottom-Up model employs Faster R-CNN \cite{ren2015faster} in conjunction with the ResNet-101 \cite{he2016deep} to generate a variably-sized set of $k$ representations, $\matr{A} = \{\vec{a_1},...,\vec{a_k}\}$, $\vec{a_i} \in R^{2048}$, such that each representation encodes a salient region of the image. We use the global average pooled image representation $\vec{a_g}=\frac{1}{k}\sum_{i}\vec{a_i}$ as our global image representation. For modeling convenience, we use a single layer of PR-FC with rectifier activation function to transform the representation vectors into new vectors with dimension $d$:
\begin{equation}
\vec{v_i} = ReLU\left( PR\text{-}FC(\matr{W_a}, \vec{a_i})\right) , \vec{v_i} \in R^d
\end{equation}
\begin{equation}
\vec{v_g} = ReLU\left( PR\text{-}FC(\matr{W_g}, \vec{a_g})\right), \vec{v_g} \in R^d
\end{equation}
where $\matr{W_a}$ and $\matr{W_g}$ are the weight parameters. The transformed $\matr{V} =\{\vec{v_1},...,\vec{v_k}\}$ is our defined regional image representations and $\vec{v_g}$ is our defined global image representation.

\begin{table*}[t]
%\begin{table}
  \centering
  \begin{threeparttable}
    %\begin{tabular}{ccccccc}
    \begin{tabular*}{\hsize}{@{}@{\extracolsep{\fill}}lcccccccccccccccc@{}}
    \toprule
    %\multirow{1}{*}{Modle}&
    %\multicolumn{8}{c}{Cross-Entropy Loss}\cr
    %\cmidrule(lr){1-9}
    Model&BLEU-1&BLEU-2&BLEU-3&BLEU-4&METEOR&ROUGE-L&CIDEr&SPICE\cr
    \midrule
    %ATT-kCC               &74.9          &58.1          &43.7          &32.6          &25.7          &-             &102.4          &-             \cr
    LSTM-A~\cite{yao2017boosting}               &73.5          &56.6          &42.9          &32.4          &25.5          &53.9          &99.8           &18.5          \cr
    SCN-LSTM$^{\Sigma}$~\cite{gan2017semantic}       &74.1          &57.8          &44.4          &34.1          &26.1          &-             &104.1          &-             \cr
    %MAT                   &73.1          &56.7          &42.9          &32.3          &25.8          &54.1          &105.8          &18.9          \cr
    Adaptive~\cite{lu2017knowing}              &74.2          &58.0          &43.9          &33.2          &26.6          &-             &108.5          &-             \cr
    %SCST:Att2in$^{\Sigma}$         &-             &-             &-             &32.8          &26.7          &55.1          &106.5          &-             \cr
    SCST:Att2all$^{\Sigma}$~\cite{rennie2017self}        &-             &-             &-             &32.2          &26.7          &54.8          &104.7          &-             \cr
    Up-Down~\cite{anderson2018bottom}               &77.2 &-             &-             &36.2          &27.0          &56.4          &113.5          &20.3          \cr
    Stack-Cap~\cite{gu2017stack}             &76.2          &60.4          &46.4          &35.2          &26.5          &-             &109.1          &-             \cr
    ARNet~\cite{chen2018regularizing}                 &74.0          &57.6          &44.0          &33.5          &26.1          &54.6          &103.4          &19.0          \cr
    NBT~\cite{lu2018neural}                   &75.5          &-             &-             &34.7          &27.1          &-             &107.2          &20.1          \cr
    %LTG-Review-Net$^{\Sigma}$ &75.1          &59.2          &45.8          &35.3          &26.7          &55.5          &107.8          &-             \cr
    GCN-LSTM$_{sem}$~\cite{yao2018exploring}  &\textbf{77.3}   &-  &-  &36.8   &27.9   &57.0   &\textbf{116.3}  &20.9   \cr
    Ours:PR &76.8&\textbf{61.0}&\textbf{47.5}&\textbf{37.0}&\textbf{28.2}&\textbf{57.1}&116.1&\textbf{21.1}   \cr
    \hline
    %SCST:Att2in$^{\Sigma*}$   &-             &-             &-             &34.8          &26.9          &56.3          &115.2          &-             \cr
    EmbeddingReward$^*$~\cite{ren2017deep}     &71.3   &53.9   &40.3   &30.4   &25.1   &52.5   &93.7   &-  \cr
    LSTM-A$^*$~\cite{yao2017boosting}       &78.6   &-  &-  &35.5   &27.3   &56.8   &118.3  &20.8   \cr
    SCST:Att2all$^{\Sigma*}$~\cite{rennie2017self}  &-             &-             &-             &35.4          &27.1          &56.6          &117.5          &-             \cr
    Up-Down$^*$~\cite{anderson2018bottom}    &79.8 &-             &-             &36.3          &27.7          &56.9          &120.1          &21.4          \cr
    Stack-Cap$^*$~\cite{gu2017stack}            &78.6          &62.5          &47.9          &36.1          &27.4          &56.9          &120.4          &20.9          \cr
    GCN-LSTM$_{sem}^*$~\cite{yao2018exploring}    &80.5   &-  &-  &38.2   &28.5   &58.3   &127.6  &22.0   \cr
    CAVP$^*$~\cite{liu2018context}     &-  &-  &-  &38.6   &28.3   &58.5   &126.3  &21.6   \cr
    SGAE$^*$~\cite{yang2018autoencoding}     &\textbf{80.8}     &-     &-  &38.4   &28.4   &\textbf{58.6}   &127.8     &22.1   \cr
    Ours:PR$^*$ &80.4&\textbf{64.9}&\textbf{50.5}&\textbf{38.7}&\textbf{28.8}&58.5&\textbf{128.3}&\textbf{22.4}\cr
    \bottomrule
    \end{tabular*}
  \end{threeparttable}
  \caption{Performance compared with the state-of-the-art methods on the Karpathy test split of MS COCO. $^{\Sigma}$ indicates ensemble. The top part is for cross-entropy training, and the bottom part is for REINFORCE-based optimization (marked with $^*$). All values are reported in percentage, with the highest value of each entry highlighted in boldface.}
  \label{tab:comparison with other methods on MS COCO Karpathy split}
\end{table*}

\noindent\textbf{Decoder.}
For decoding image representations $\bm{V}$ and $\bm{v}_g$ to sentence description, as shown in Figure \ref{fig:caption_decoder}, we utilize an visual attention model with two PR-LSTM layers according to recent methods \cite{anderson2018bottom,lu2018neural,yao2018exploring}, which are characterized as Attention PR-LSTM and Language PR-LSTM respectively. We initialize the hidden state and memory cell of each PR-LSTM as zero.
\begin{figure}[t]
    \centering
    \includegraphics[width=\columnwidth]{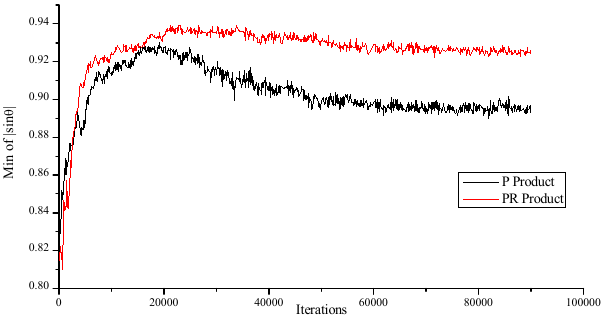}
    \caption{The minimum of $|\sin \theta|$ of the hidden-hidden transfer part in the Attention LSTM.} %Compared with the P Product version (black), the PR Product version (red) behaves very differently.}
    \label{fig:a1-min_a2-max}
\end{figure}

Given the output $\vec{h_t^1}$ of the Attention PR-LSTM, we generate the attended regional image representation $\vec{\hat{v}_t}$ through the attention model, which is broadly adopted in recent previous work \cite{chen2017sca,lu2017knowing,anderson2018bottom}. Here, we use the PR Product version of visual attention model expressed as follows:
\begin{equation}
\begin{split}
\vec{f_1}&=tanh\left(PR\text{-}FC(\matr{W_v}, \matr{V})+PR\text{-}FC(\matr{W_{h1}}, \vec{h_t^1})\right)\\
\vec{f_2}&=PR\text{-}FC(\matr{W_z},\vec{f_1})\\
\vec{\alpha_t}&=softmax\left(\vec{f_2}\right)\\
\vec{\hat{v}_t}&=\sum_{i=1}^{k}\alpha_{t,i}\vec{v_i}
\end{split}
\end{equation}
where $\matr{W_v}, \matr{W_{h1}}$ and $\matr{W_z}$ are learned parameters, $\vec{f_1}$ and $\vec{f_2}$ are the outputs of the first layer and the second layer in the attention model respectively. $\vec{\alpha_t}$ is the attention weight over $k$ regional image representations, and $\vec{\hat{v}_t}$ is the attended image representation at time step t.

\subsection{Dataset and Settings}
\noindent \textbf{Dataset.}
We evaluate our proposed method on the MS COCO dataset \cite{lin2014microsoft}. MS COCO dataset contains 123287 images labeled with at least 5 captions. There are 82783 training images and 40504 validation images, and it provides 40775 images as the test set for online evaluation as well. For offline evaluation, we use a set of 5000 images for validation, a set of 5000 images for test and the remains for training, as given in \cite{karpathy2015deep}. We truncate captions longer than 16 words and then build a vocabulary of words that occur at least 5 times in the training set, resulting in 9487 words.

%\noindent \textbf{Evaluation Metrics.}
%We report results using the COCO captioning evaluation toolkit \cite{lin2014microsoft}, which reports the widely used automatic evaluation metrics: BLEU(including BLEU-1, BLEU-2, BLEU-3, BLEU-4), METEOR, ROUGE-L, CIDEr and SPICE.

\noindent \textbf{Implementation Details.}
In the captioning model, we set the number of hidden units in each LSTM or PR-LSTM to 512, the embedding dimension of a word to 512, and the embedding dimension of image representation to 512. All of our models are trained according to the following recipe. We train all models under the cross-entropy loss using ADAM optimizer with an initial learning rate of $5 \times 10^{-4}$ and a momentum parameter of 0.9. We anneal the learning rate using cosine decay schedule and increase the probability of feeding back a sample of the word posterior by 0.05 every 5 epochs until we reach a feedback probability 0.25 \cite{bengio2015scheduled}.
%We evaluate the model at every 6000 iterations on the validation set and select the last evaluated model as initialization for REINFORCE training.
We then run REINFORCE training to optimize the CIDEr metric using ADAM with a learning rate $5 \times 10^{-5}$ with cosine decay schedule and a momentum parameter of 0.9. During CIDEr optimization mode and testing mode, we use a beam size of 5. Note that in all our model variants, the untransformed image representations $\matr{A}$ and $\vec{a_g}$ from the Encoder are fixed and not fine-tuned. As our focus is on the effectiveness of our proposed PR Product, so we just exploit the widely used backbone model and settings, without any additional tricks of improving the performance, like scene graph and ensemble strategy.

\subsection{Performance Comparison and Experimental Analysis}
\noindent \textbf{The effectiveness of PR Product.}
To test the effectiveness of PR Product, we first compare the performance of models using the following different substitutes for inner product on Karpathy's split of MS COCO dataset:
\begin{itemize}
  \item \textbf{P Product}: This is just the standard inner product. %, which only involves the information of vector projection of $\vec{x}$ on $\vec{w}$, as shown in Equation (\ref{equ:innerproduct_geo}).
      In Euclidean geometry, it is also called projection product, so we abbreviate it as P Product.
  \item \textbf{R Product}: Contrary to P Product, R Product only involves the information of vector rejection of $\vec{x}$ from $\vec{w}$. To keep the same range and sign as P Product, we formulate the R Product as follows:
      \begin{equation}\label{equ:R_Net}
        R(\vec{w},\vec{x})=sign(\cos\theta)\|\vec{w}\|_2(\|\vec{x}\|_2-\|\Vec{R_x}\|_2)
      \end{equation}
  \item \textbf{PR Product}: This is the proposed PR Product. %, which involves not only the information of vector projection $\Vec{P_x}$ but also the one of vector rejection $\Vec{R_x}$, as shown in Equation (\ref{equ:PR-Net}).
      Evidently, the PR Product is the combination of the P Product and R Product with the relationship as follows:
      \begin{equation}\label{equ:relation_P_R_PR}
        PR(\vec{w},\vec{x})=\underline{|\sin \theta|}P(\vec{w},\vec{x}) + \underline{|\cos \theta|}R(\vec{w},\vec{x})
      \end{equation}
\end{itemize}
\begin{table*}[!t]
%\begin{table}
  \centering
  \begin{threeparttable}
    %\begin{tabular}{ccccccc}
    \begin{tabular*}{\hsize}{@{}@{\extracolsep{\fill}}p{2.5cm}cccccccccccccccc@{}}
    \toprule
    \multirow{2}{*}{}&
    \multicolumn{2}{c}{BLEU-1}&\multicolumn{2}{c}{BLEU-2}&\multicolumn{2}{c}{BLEU-3}&\multicolumn{2}{c}{BLEU-4}&\multicolumn{2}{c}{METEOR}&\multicolumn{2}{c}{ROUGE-L}&\multicolumn{2}{c}{CIDEr}\cr
    \cmidrule(lr){2-3} \cmidrule(lr){4-5} \cmidrule(lr){6-7} \cmidrule(lr){8-9} \cmidrule(lr){10-11} \cmidrule(lr){12-13} \cmidrule(lr){14-15}
    &C5&C40&C5&C40&C5&C40&C5&C40&C5&C40&C5&C40&C5&C40\cr
    \midrule
    SCN-LSTM$^{\Sigma}$~\cite{gan2017semantic}       &74.0 &91.7 &57.5 &83.9 &43.6 &73.9 &33.1 &63.1 &25.7 &34.8 &54.3 &69.6 &100.3 &101.3 \cr
    Adaptive$^{\Sigma}$~\cite{lu2017knowing}      &74.8 &92.0 &58.4 &84.5 &44.4 &74.4 &33.6 &63.7 &26.4 &35.9 &55.0 &70.5 &104.2 &105.9 \cr
    SCST:Att2all$^{*\Sigma}$~\cite{rennie2017self}  &78.1 &93.7 &61.9 &86.0 &47.0 &75.9 &35.2 &64.5 &27.0 &35.5 &56.3 &70.7 &114.7 &116.7 \cr
    Up-Down$^{*\Sigma}$~\cite{anderson2018bottom}       &\textbf{80.2} &\textbf{95.2} &64.1 &\textbf{88.8} &49.1 &\textbf{79.4} &36.9 &\textbf{68.5} &27.6 &36.7 &57.1 &72.4 &117.9 &120.5 \cr
    %\small GCN-LSTM$^{*\Sigma}$\cite{yao2018exploring}    &-  &-  &65.5   &89.3   &50.8    &80.3   &38.7   &69.7   &28.5   &37.6   &58.5   &73.4   &125.3  &126.5  \cr
    %SGAE$^{*\Sigma}$~\cite{yang2018autoencoding}   &81.0   &95.3   &65.6   &89.5   &50.7   &80.4   &38.5   &69.7   &28.2   &37.2   &58.6   &73.6   &123.8  &126.5  \cr
    %\hline
    %ATT-kCC               &74.3 &91.5 &57.5 &83.2 &43.1 &72.2 &32.1 &60.7 &25.5 &34.1 &-    &-    &98.7  &100.1 \cr
    LSTM-A~\cite{yao2017boosting}               &78.7 &93.7 &62.7 &86.7 &47.6 &76.5 &35.6 &65.2 &27.0 &35.4 &56.4 &70.5 &116.0   &118.0   \cr
    PG-BCMR$^*$~\cite{liu2017improved}              &75.4 &91.8 &59.1 &84.1 &44.5 &73.8 &33.2 &62.4 &25.7 &34.0 &55.0 &69.5 &101.3 &103.2 \cr
    MAT~\cite{liu2017mat}                   &73.4 &91.1 &56.8 &83.1 &42.7 &72.7 &32.0 &61.7 &25.8 &34.8 &54.0 &69.1 &102.9 &106.4 \cr
    Stack-Cap$^*$~\cite{gu2017stack}            &77.8 &93.2 &61.6 &86.1 &46.8 &76.0 &34.9 &64.6 &27.0 &35.6 &56.2 &70.6 &114.8 &118.3 \cr
    %CAVP$^*$~\cite{liu2018context}   &80.1    &94.9   &64.7   &88.8   &50.0   &79.7   &37.9   &69.0   &28.1   &37.0   &58.2   &73.1   &121.6  &123.8  \cr
    %SGAE$^*$~\cite{yang2018autoencoding}    &80.6   &95.0   &65.0   &88.9   &50.1   &79.6   &37.8   &68.7   &28.1   &37.0   &58.2   &73.1   &122.7  &125.5  \cr
    Ours:PR$^*$              &79.9 &94.5	      &\textbf{64.3} &88.2          &\textbf{49.6} &79.0 &\textbf{37.7} &68.3 &\textbf{28.4} &\textbf{37.5} &\textbf{58.0} &\textbf{73.0} &\textbf{122.3} &\textbf{124.1} \cr
    \bottomrule
    \end{tabular*}
  \end{threeparttable}
  \caption{Performance compared with the state-of-the-art methods on the online MS COCO test server. $^{\Sigma}$ indicates ensemble, and $^*$ indicates fine-tuned by REINFORCE-based optimization. The top part is for the ensemble models, and the bottom part is for the singles. All values are reported in percentage, with the highest value of each entry highlighted in boldface.}
  \label{tab:comparison with other methods on MS COCO test server}
\end{table*}

For fair comparison, results are reported for models trained with cross-entropy loss and models optimized for CIDEr score on Karpathy's split of MS COCO dataset, as shown in Table \ref{tab:model variants_comparison}. Although the R Product does not perform as well as the P Product or PR Product, the results show that the vector rejection of data vector from weight vector can be used to optimize neural networks. Compared with the P Product and R Product, the PR Product achieves performance improvement across all metrics regardless of cross-entropy training or CIDEr optimization, which experimentally proves the cooperation of vector projection and vector rejection is beneficial to the optimization of neural networks. To intuitively illustrate the advantage of the PR Product, we show some examples of image captioning in supplementary material.

To better understand how the PR Product affects neural networks, we plot the minimum of $|\sin \theta|$ to investigate the dynamics of neural networks to some extent. Figure \ref{fig:a1-min_a2-max} shows the statistic of the hidden-hidden transfer part in the Attention LSTM, and plots for more layers can be found in the supplementary material.
%two LSTMs, the attention module and the output layer.
For most of the layers, the minimum of $|\sin \theta|$ in PR Product version is larger than the one in P Product, which means the weight vector and data vector in PR Product are more orthogonal. We argue this is the reason for PR Product to take effect.

\noindent \textbf{Comparison with State-of-the-Art Methods.}
To further verify the effectiveness of our proposed method, we also compare the PR Product version of our captioning model with some state-of-the-art methods on Karpathy's split of MS COCO dataset. Results are reported in Table \ref{tab:comparison with other methods on MS COCO Karpathy split}, of which the top part is for cross-entropy loss and the bottom part is for CIDEr optimization.
% which include ATT-kCC in \cite{mun2017text} , LSTM-A3 in \cite{yao2017boosting}, SCN-LSTM in \cite{gan2017semantic}, MAT in \cite{liu2017mat}, Adaptive in \cite{lu2017knowing}, SCST:Att2in in \cite{rennie2017self}, SCST:Att2all in \cite{rennie2017self}, Up-Down in \cite{anderson2018bottom}, Stack-Cap in \cite{gu2017stack} , ARNet in \cite{chen2018regularizing}, NBT in \cite{lu2018neural}, LTG-Review-Net in \cite{jiang2018learning} and PG-BCMR in \cite{liu2017improved}.

Among those methods, SCN-LSTM~\cite{gan2017semantic} and SCST:Att2all~\cite{rennie2017self} use the ensemble strategy. GCN-LSTM~\cite{yao2018exploring}, CAVP~\cite{liu2018context} and SGAE~\cite{yang2018autoencoding} exploit information of visual scene graphs. Even though we do not use any of the above means of improving performance, our PR Product version of captioning model achieves the best performance in most of the metrics, regardless of cross-entropy training or CIDEr optimization.
In addition, we also report our results on the official MS COCO evaluation server in Table \ref{tab:comparison with other methods on MS COCO test server}. As the scene graph models can greatly improve the performance, for fair comparison, we only report the results of methods without scene graph models. It is noteworthy that we just use the same model as reported in Table \ref{tab:comparison with other methods on MS COCO Karpathy split}, without retraining on the whole training and validation images of MS COCO dataset.
We can see that our single model achieves competitive performance compared with the state-of-the-art models, even though some models exploit ensemble strategy.
%\noindent \textbf{Qualitative Analysis}
%We show some representative examples in Figure \ref{fig:caption_example} to illustrate the advantage of our proposed Language Attention module. The images are selected from Karpathy's test split of MS COCO dataset. Both the \emph{VA} baseline model and our proposed \emph{LA$_l$} model are trained with cross-entropy loss and then fine-tuned for CIDEr optimization. The results show that using the Language Attention module makes contribution to the descriptiveness of the sentences. For example, in image 2, image 3 and image 7, the sentences generated by the \emph{VA} baseline model contain duplicate words, which makes the meaning confused, while our \emph{LA$_l$} model generates meaningful sentences. And in the sentences for image 4, image 5 and image 8, our \emph{LA$_l$} model chooses more appropriate words. Specifically in image 8, the \emph{VA} model uses `sitting' after `a green car', and our model uses a more appropriate `parked'.
%Even more amazing is that in image 2, the Language Attention module actually assists the caption model to notice the `bird' which just takes up a little bit of area in the image, but the \emph{VA} baseline model does not. These examples demonstrate that our proposed Language Attention module can regularize the sentences to a certain extent, which makes the sentence grammar more correct and semantics more accurate.

\section{Conclusion}
In this paper, we propose a reliable substitute for the inner product of weight vector $\vec{w}$ and data vector $\vec{x}$, the PR Product, which involves the information of both the vector projection $\Vec{P_x}$ and the vector rejection $\Vec{R_x}$. The length of the direction gradient of PR Product w.r.t. $\vec{w}$ is consistently larger than the one in standard inner product. In particular, we show the PR Product version of the fully connected layer, convolutional layer and LSTM layer. Applying these PR Product version modules to image classification and image captioning, the results demonstrate the robust effectiveness of our proposed PR Product. As the basic operation in neural networks, we will apply the PR Product to other tasks like object detection.

~\\
\noindent \textbf{Acknowledgement.}
This work was supported in part by the NSFC Project under Grants 61771321 and 61872429, in part by the Guangdong Key Research Platform of Universities under Grants 2018WCXTD015, in part by the Science and Technology Program of Shenzhen under Grants KQJSCX20170327151357330, JCYJ20170818091621856, and JSGG20170822153717702, and in part by the Interdisciplinary Innovation Team of Shenzhen University.

{\small
\bibliographystyle{ieee_fullname}
\bibliography{egbib}
}
%
%\clearpage
%\section{Supplementary material for ICCV submission 5532}
%\subsection{The Minimum of $\pmb{|\sin\theta|}$ and the Maximum of $\pmb{|\cos\theta|}$}
%We plot the minimum of $|\sin \theta|$ and the maximum of $|\cos \theta|$ of some layers in the following, from which we can see that the PR Product version and the P Product version behave very differently.

%\end{document}

\newpage
%%%%%%%%%% TITLE
%\title{Supplementary material for\\
%PR Product: A Substitute for Inner Product in Neural Networks}
%\maketitle
%%\vspace*{-25mm}
%\thispagestyle{empty}

\appendix

\section{Training Curves on CIFAR10}
\begin{figure}[H]
    \centering
    \includegraphics[width=0.9\columnwidth, height=0.48\columnwidth]{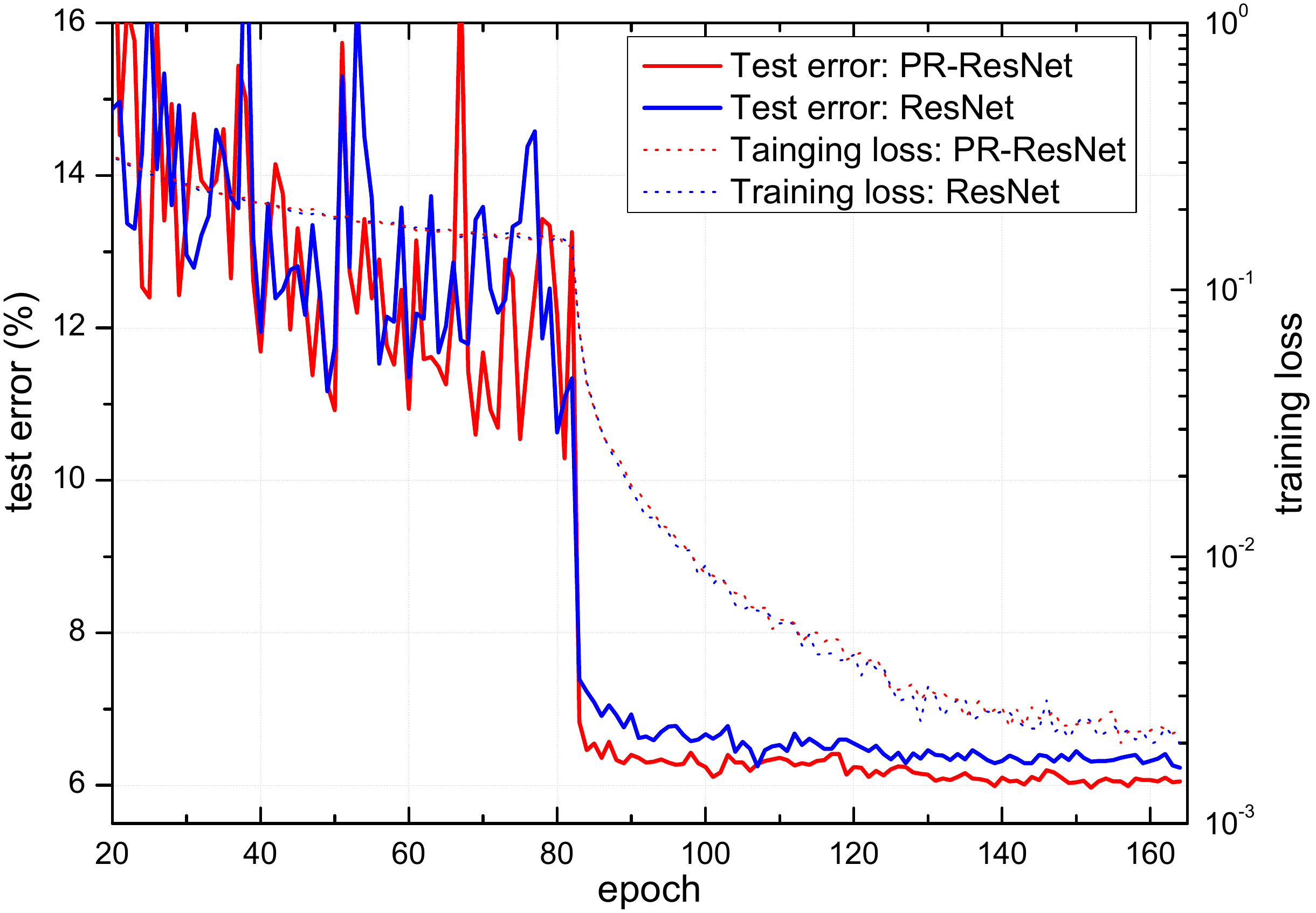}
    \caption{Training curves of the ResNet on CIFAR10.}
    \label{fig:cifar10_resnet110}
\end{figure}

\begin{figure}[H]
    \centering
    \includegraphics[width=0.9\columnwidth, height=0.48\columnwidth]{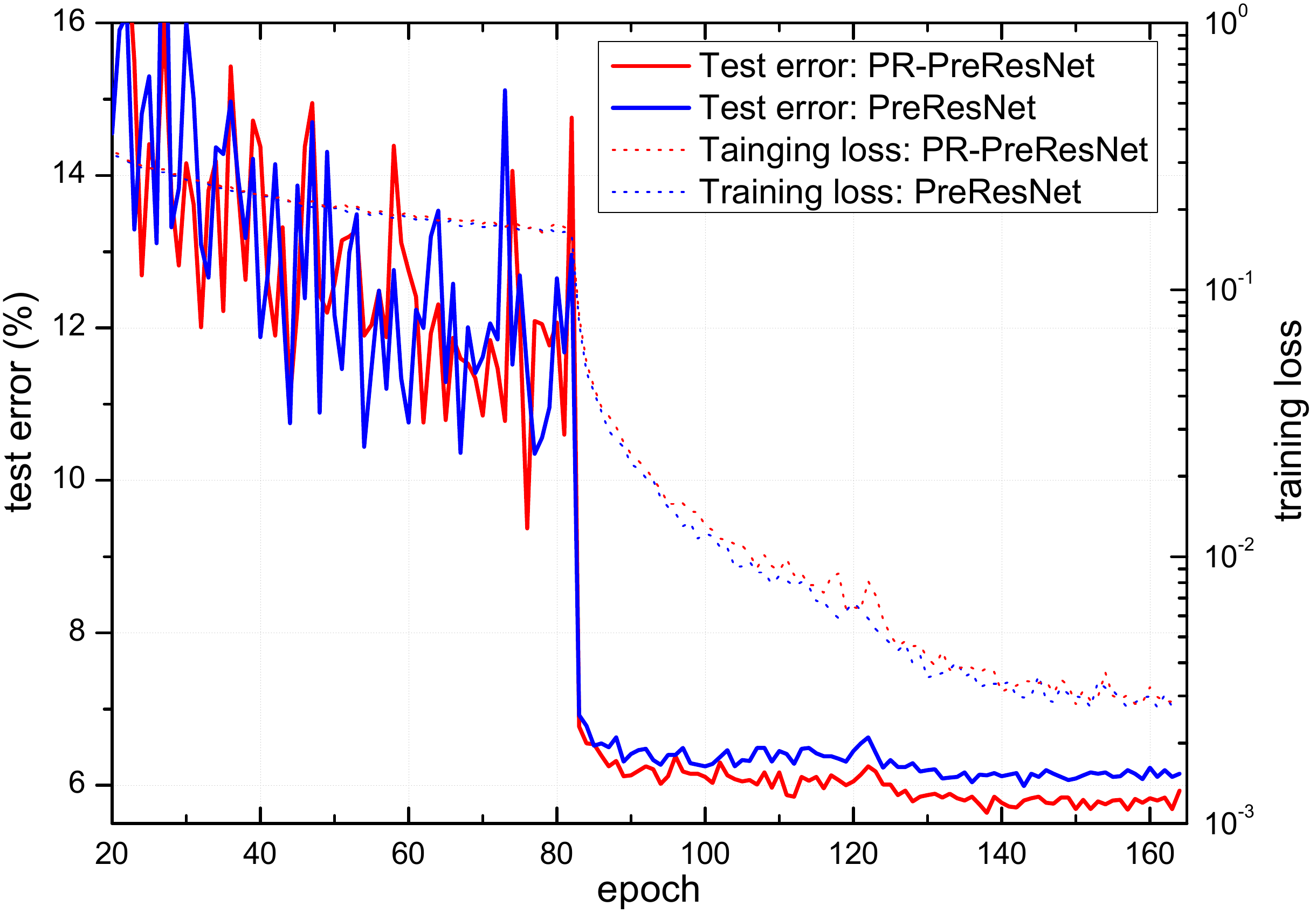}
    \caption{Training curves of the PreResNet on CIFAR10.}
    \label{fig:cifar10_preresnet110}
\end{figure}

\begin{figure}[H]
    \centering
    \includegraphics[width=0.9\columnwidth, height=0.48\columnwidth]{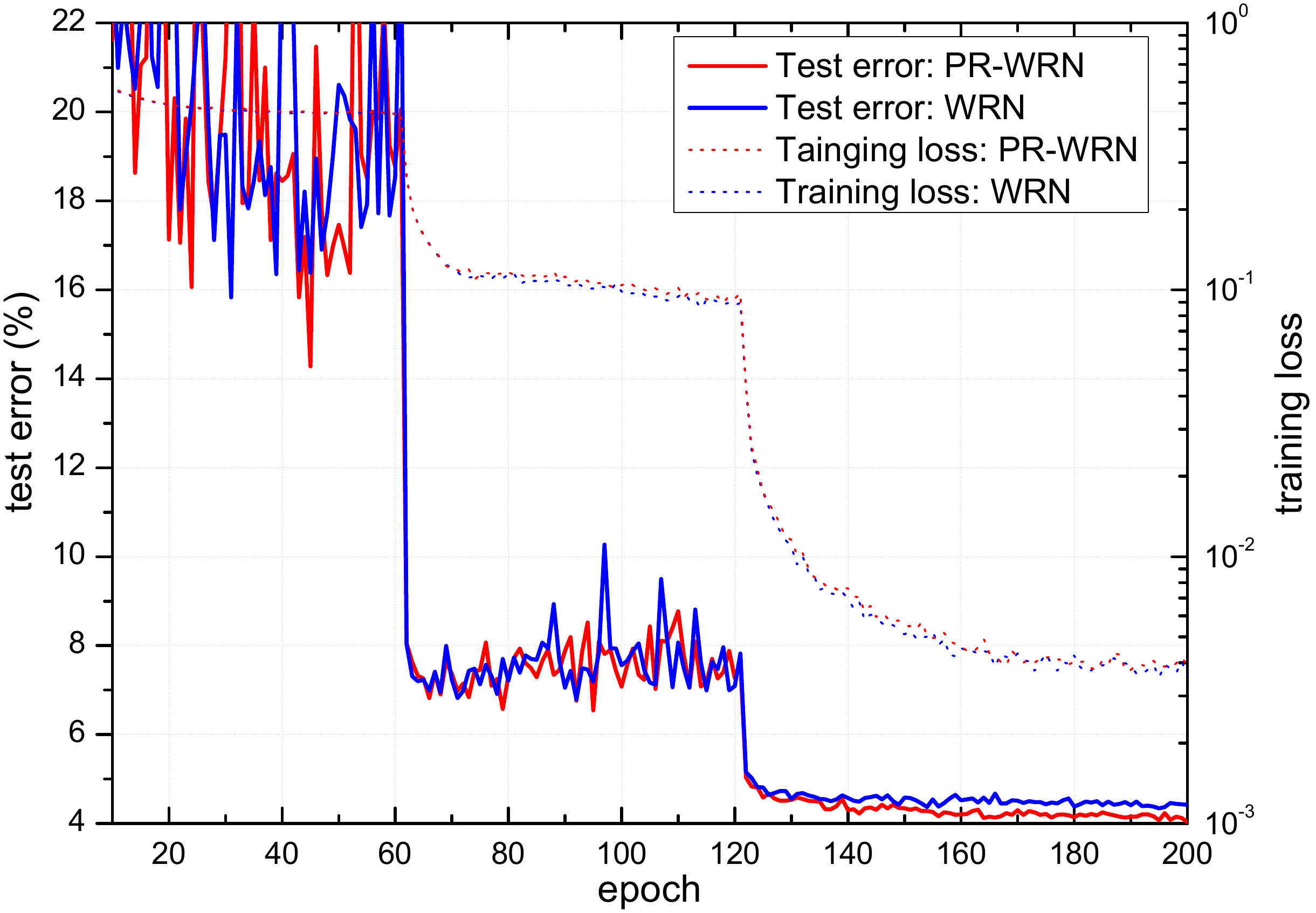}
    \caption{Training curves of the WRN on CIFAR10.}
    \label{fig:cifar10_WRN}
\end{figure}

\begin{figure}[H]
    \centering
    \includegraphics[width=0.9\columnwidth, height=0.48\columnwidth]{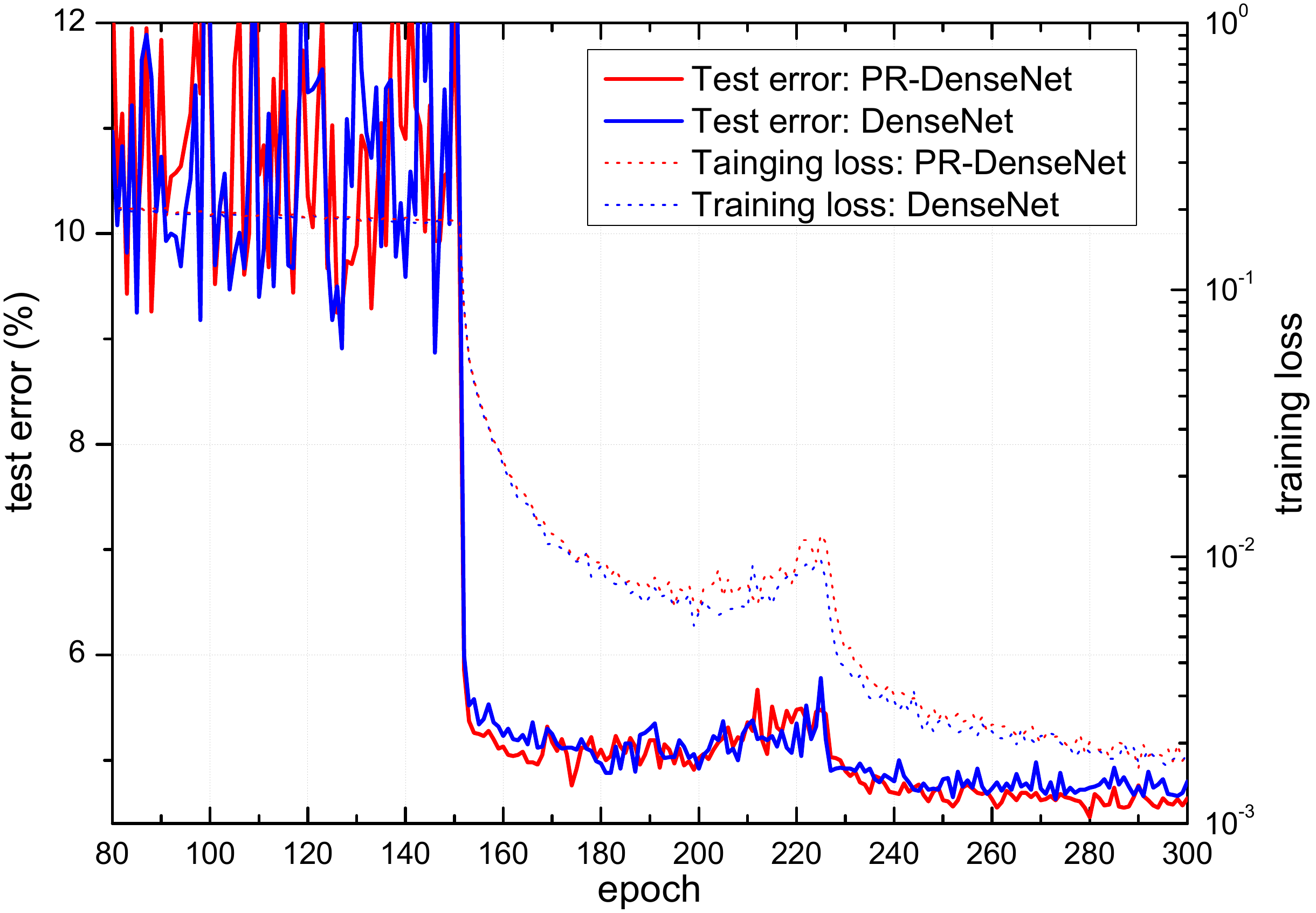}
    \caption{Training curves of the DenseNet on CIFAR10.}
    \label{fig:cifar10_DenseNet}
\end{figure}

Figures \ref{fig:cifar10_resnet110}-\ref{fig:cifar10_DenseNet} show the training curves of some classification models on CIFAR10 used in the paper,  from which we can see that the models of PR Product version get consistent lower error rates than the models of standard inner product version.

\section{The Minimum of $\pmb{|\sin\theta|}$}% and the Maximum of $\pmb{|\cos\theta|}$}
\begin{figure}[H]
    \centering
    \includegraphics[width=0.9\columnwidth]{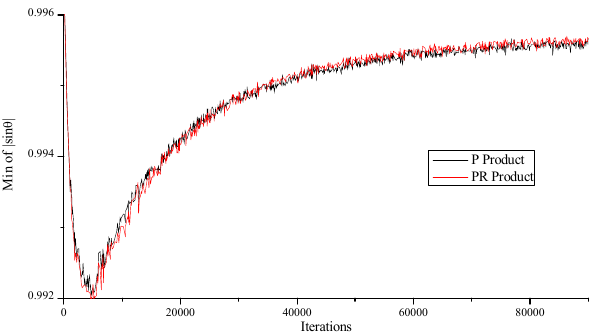}
    \caption{The minimum of $|\sin \theta|$ of the $\vec{a_i}$ to $\vec{v_i}$ transfer part in the Encoder.}
    \label{fig:vi}
\end{figure}

\begin{figure}[H]
    \centering
    \includegraphics[width=0.9\columnwidth]{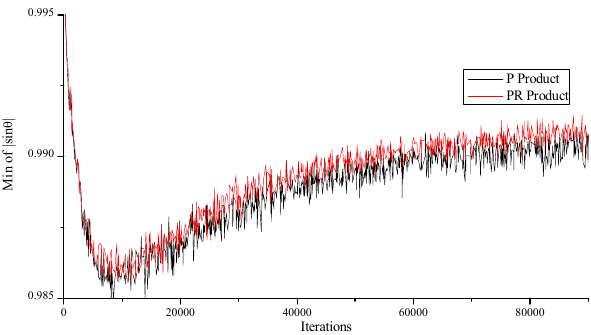}
    \caption{The minimum of $|\sin \theta|$ of the $\vec{a_g}$ to $\vec{v_g}$ transfer part in the Encoder.}
    \label{fig:vg}
\end{figure}

\begin{figure}[H]
    \centering
    \includegraphics[width=0.9\columnwidth]{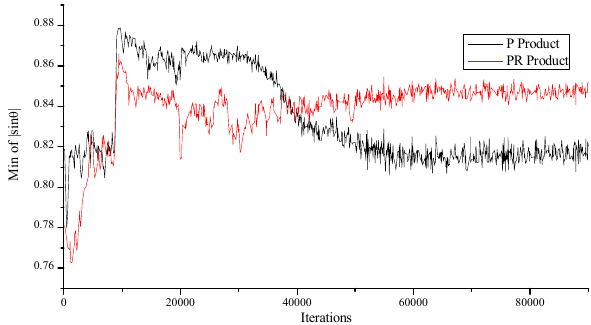}
    \caption{The minimum of $|\sin \theta|$ of the $\matr{W_e}\vec{\Pi_t}$ to hidden transfer part in the Attention LSTM.}
    \label{fig:att-lstm-word-hidden}
\end{figure}

\begin{figure}[H]
    \centering
    \includegraphics[width=0.9\columnwidth]{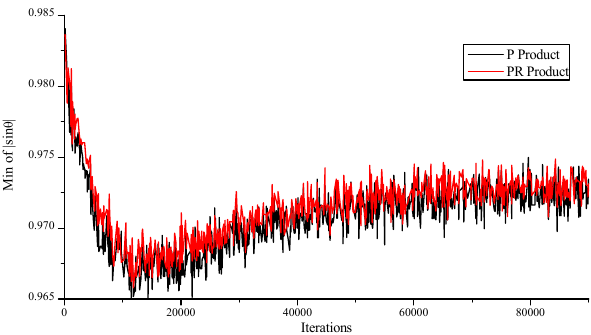}
    \caption{The minimum of $|\sin \theta|$ of the $\vec{v_g}$ to hidden transfer part in the Attention LSTM.}
    \label{fig:att-lstm-vgh}
\end{figure}

\begin{figure}[H]
    \centering
    \includegraphics[width=0.9\columnwidth]{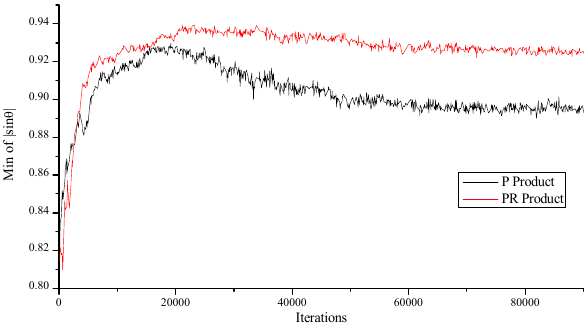}
    \caption{The minimum of $|\sin \theta|$ of the hidden to hidden transfer part in the Attention LSTM.}
    \label{fig:att-lstm-hh}
\end{figure}

\begin{figure}[H]
    \centering
    \includegraphics[width=0.9\columnwidth]{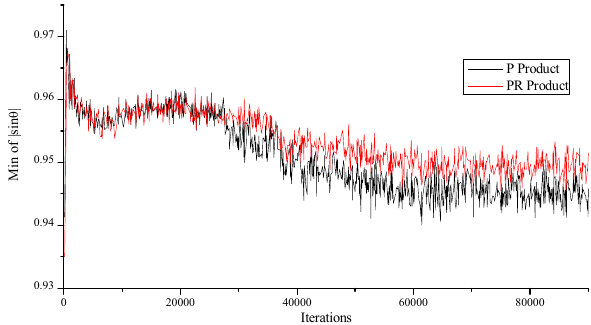}
    \caption{The minimum of $|\sin \theta|$ of the $\vec{\hat{v}_t}$ to hidden transfer part in the Language LSTM.}
    \label{fig:lang-lstm-vt-hidden}
\end{figure}

\begin{figure}[H]
    \centering
    \includegraphics[width=0.9\columnwidth]{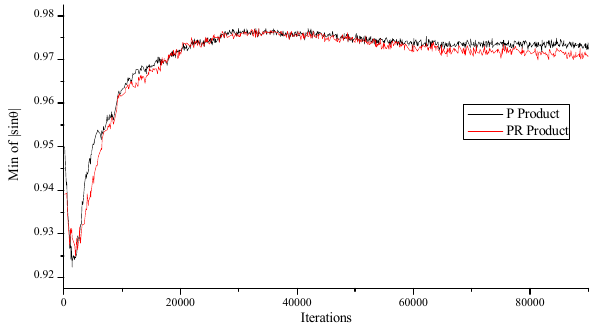}
    \caption{The minimum of $|\sin \theta|$ of the $\vec{h_t^1}$ to hidden transfer part in the Language LSTM.}
    \label{fig:lang-lstm-h1-h}
\end{figure}

\begin{figure}[H]
    \centering
    \includegraphics[width=0.9\columnwidth]{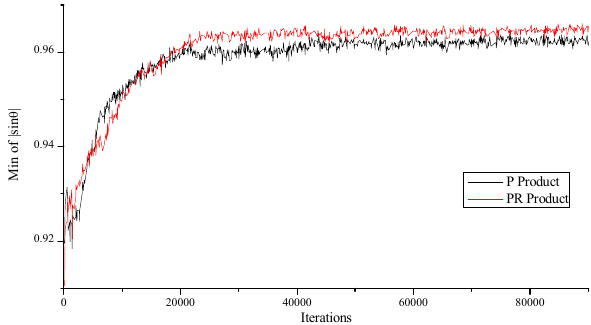}
    \caption{The minimum of $|\sin \theta|$ of the hidden to hidden transfer part in the Language LSTM.}
    \label{fig:lang-lstm-hh}
\end{figure}

\begin{figure}[H]
    \centering
    \includegraphics[width=0.9\columnwidth]{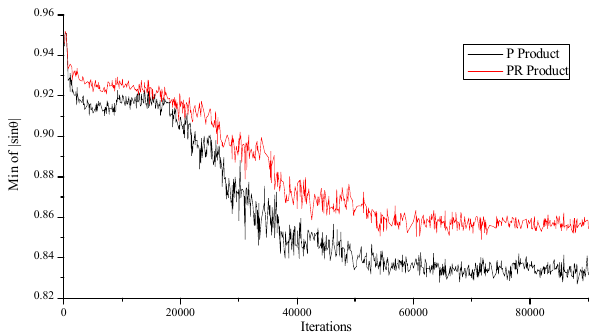}
    \caption{The minimum of $|\sin \theta|$ of the output layer (softmax layer in the Decoder of our captioning model.)}
    \label{fig:logit}
\end{figure}

We plot the minimum of $|\sin \theta|$ in some layers of our captioning model, as shown in Figures \ref{fig:vi}-\ref{fig:logit}. From these plots, we can observe that the minimum of $|\sin \theta|$ in PR Product version is larger than the one in P Product version for most of the layers, which means the weight vector and data vector in PR Product are more orthogonal. We argue this is the reason for PR Product to take effect.

\section{Examples of Image Captioning}
\begin{figure*}[ht]
    \centering
    \includegraphics[width=\textwidth]{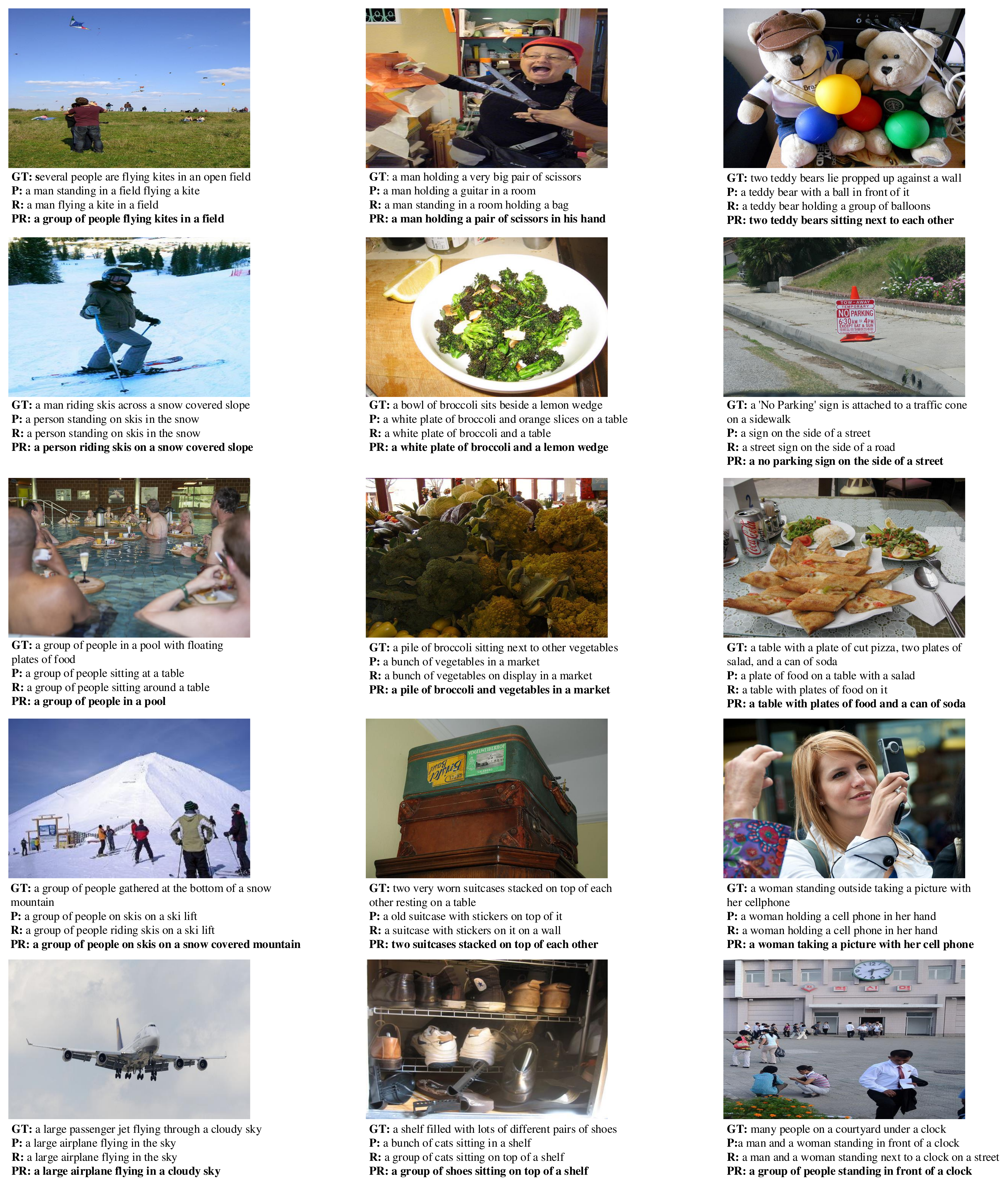}
    \caption{Examples of captions on MS COCO dataset. GT: human ground truth. P: sentence generated by the P Product version model. R: sentence generated by the R Product version model. PR: sentence generated by the PR Product version model. Obviously, the PR Product performs better than the P Product and the R Product.}
    \label{fig:caption_examples}
\end{figure*}

To intuitively illustrate the advantage of the PR Product, we show some examples of image captioning in Figure \ref{fig:caption_examples}. The images are sampled from Karpathy's test split of MS COCO dataset. All the three models (P version, R version, and PR version) are trained with cross-entropy loss and then fine-tuned for CIDEr optimization. The results show that PR product makes contribution to the descriptiveness of the sentences and prove that the PR Product is effective.

\end{document}